%% file: neurips_2026.tex
\documentclass{article}

\PassOptionsToPackage{numbers,sort&compress}{natbib}

\usepackage[preprint]{neurips_2026}

\usepackage[utf8]{inputenc} 
\usepackage[T1]{fontenc}    
\usepackage{hyperref}       
\usepackage{url}            
\usepackage{booktabs}       
\usepackage{amsfonts}       
\usepackage{amsmath}        
\usepackage{nicefrac}       
\usepackage{microtype}      
\usepackage{xcolor}         
\usepackage{graphicx}       
\usepackage{subcaption}     
\usepackage{listings}       

\title{Rethinking Scribble-Guided Image Editing: Generalization, Instruction Adherence, and Multi-Tasking}

\author{%
  Mingyi Xu$^{1,2}$\thanks{Work done when Mingyi Xu was an intern at Taobao \& Tmall Group of Alibaba.}
  \quad
  Jinpeng Lin$^{2}$
  \quad
  Min Zhou$^{2}$
  \quad
  Tiezheng Ge$^{2}$
  \quad
  Ming Zeng$^{1}$ \\
  $^{1}$Xiamen University
  \quad
  $^{2}$Taobao \& Tmall Group of Alibaba
}

\begin{document}

\maketitle

\input{sections/0_abstract.tex}
\input{sections/1_introduction.tex}

\input{sections/2_related_works.tex}

\input{sections/3_empirical_study.tex}
\input{sections/4_methods.tex}

\input{sections/5_experiments.tex}
\input{sections/6_conclusion.tex}

\section*{References}
{
\small
\bibliographystyle{plainnat}
\bibliography{reference}
}

\appendix
\input{sections/X_supply}


\end{document}

%% file: sections/0_abstract.tex
\begin{abstract}
Scribble-guided image editing is an image editing paradigm in which users combine simple scribble annotations with text prompts to specify both where and how an image should be edited, enabling more flexible interaction while providing clearer editing guidance and more precise spatial control. However, existing image editing models still exhibit unstable performance under this paradigm, and the challenge becomes even greater in multi-task scenarios.
To improve performance in scribble-guided image editing, we conducted empirical studies using an open-source editing model. Our experiments reveal an asymmetry in generalization: instruction-level generalization, including across editing tasks and from single-task to multi-task settings, is more challenging than image-domain generalization, such as from synthetic to real-world images or from mosaicked to regular images. This suggests that the primary bottleneck lies not in the image domain gap, but in insufficient learning for diverse editing instructions. Motivated by this insight, we propose three simple yet effective strategies: (a) a Coverage-then-Realism Curriculum with a two-stage data construction and training pipeline, where we first build large-scale synthetic, instruction-rich data to provide broad supervision over editing tasks, and then curate a small amount of real-world image editing data to refine generation realism; (b) Multi-Task Mosaicking, which constructs multi-task training samples by concatenating single-task examples at nearly zero cost while enabling the learned multi-task editing capability to generalize to non-mosaicked images; and (c) an Edit-Focused Loss, which leverages the changed regions between input and output images available in synthetic data to focus training on edited regions, thereby improving both learning efficiency and editing accuracy. With these strategies, we substantially improve the performance of an open-source model on both single-task and multi-task scribble-guided editing on VIBE benchmark, achieving state-of-the-art results. We will publicly release our dataset and model.

\end{abstract}

%% file: sections/1_introduction.tex
\section{Introduction}

\begin{figure}[t]
  \centering
  \includegraphics[width=0.95\linewidth]{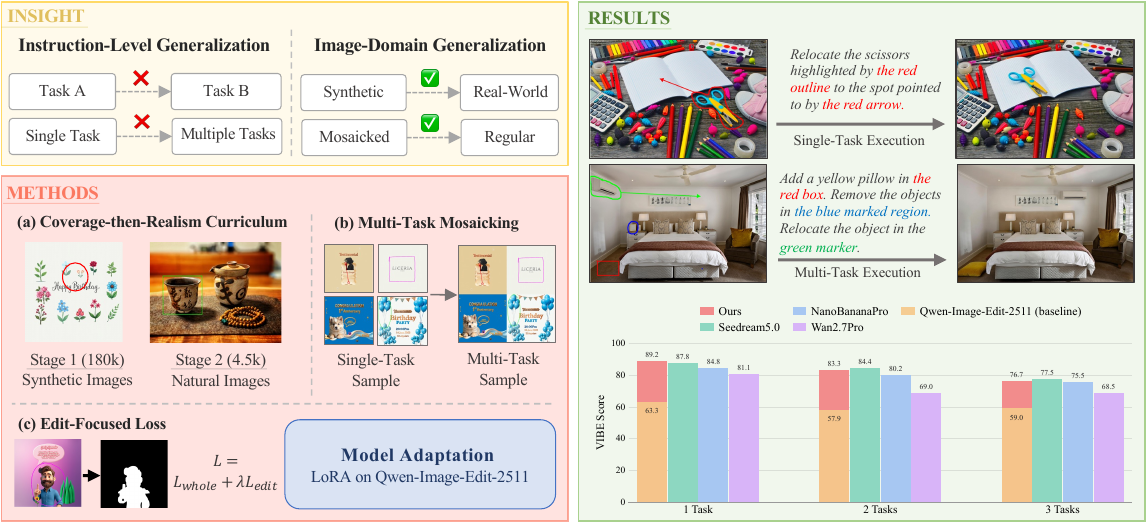}

  \caption{Overview. Two empirical observations on scribble-guided image editing motivate the three strategies we introduce (left),
  which together let an open-source backbone (Qwen-Image-Edit-2511) achieve state-of-the-art
  results on VIBE (right).}
  \label{fig:teaser}
\end{figure}
\vspace{-0.3cm}
  
Text-guided image editing has made strong progress in recent years~\citep{shuai2024survey,brooks2023instructpix2pix}. However, text prompts alone often fail to specify where edits should be applied. Scribble-guided image editing addresses this issue by allowing users to provide a few coarse strokes on images. Despite this simple and flexible interaction, the task remains challenging for current models, especially when multiple objects must be edited simultaneously. The recent benchmark VIBE~\citep{zhang2026vibe} further shows that open-source models still lag behind closed-source systems in this setting.

To investigate this problem and explore how to narrow the performance gap between open-source and closed-source models, we conduct a set of empirical studies using the open-source editing model Qwen-Image-Edit-2511~\citep{wu2025qwenimage}. Our experiments reveal a clear pattern: the model struggles much more with instruction-level generalization than with image-domain generalization. Specifically, a model trained on the ``object removal” task achieves a score of 87.62 for the same task on VIBE, but only 16.99 on the ``object translation” task. In contrast, a model trained purely on synthetic data can still perform correct edits on real-world images, although the realism of the generated results is somewhat weaker. We further find that the performance drop from single-task to multi-task settings is not mainly caused by the image-domain shift, i.e., the binding ambiguity~\citep{DBLP:conf/iccv/XieLHLZ0S23} introduced when moving from one scribble to multiple scribbles. Simply adding distractor scribbles to single-task training images does not noticeably improve multi-task editing. Instead, the key bottleneck is insufficient learning of complete tuples $\langle \text{multi-scribble image}, \text{multi-instruction prompt}, \text{multi-edit ground truth} \rangle$. Training on multi-task data constructed by mosaicking multiple single-task image-text pairs significantly improves multi-task editing, and this capability generalizes well to regular non-mosaicked images.

Guided by this insight, we design three strategies shown in Figure~\ref{fig:teaser} for scribble-guided image editing.

First, we propose a \textbf{Coverage-then-Realism Curriculum}, a two-stage data construction and training framework. Based on an open-source layered image dataset~\citep{chen2025prismlayers}, we use a vision-language model (VLM)~\citep{shuai2025qwen3vl} to select objects and generate task-specific editing prompts and bounding boxes~(scribbles), resulting in 180K synthetic training samples. We further collect real-world internet images as target images, use a VLM to generate task-specific editing instructions, apply Nano Banana Pro~\citep{deepmind2025nanobananapro} to construct corresponding input images, and obtain bounding boxes~(scribbles) with a segmentation model. After manual filtering, this process yields 4.5K training samples. In the first training stage, we use the synthetic data covering diverse editing tasks to build broad editing and instruction-following capabilities. In the second stage, we utilize the small amount of real-image editing data to improve alignment with real-world image distributions and enhance output realism.

Second, we introduce \textbf{Multi-Task Mosaicking}, a simple yet effective strategy for constructing multi-task training samples by concatenating multiple single-task examples with almost no extra annotation or collection cost. Although the resulting training samples are mosaicked compositions rather than regular images, we find that the learned capability transfers well to non-mosaicked multi-task images, yielding clear improvements on VIBE’s 2-task and 3-task evaluations.

Third, when synthetic data are available, the edited regions between the input image and the target image can be obtained automatically and accurately. We therefore incorporate an \textbf{Edit-Focused Loss} that places additional training emphasis on these regions of change. By explicitly focusing optimization on the parts of the image that are most relevant to the requested edits, this objective improves both learning efficiency and editing precision.

As shown in Figure~\ref{fig:teaser}, these strategies substantially improve both single-task and multi-task scribble-guided editing, enabling Qwen-Image-Edit-2511 to achieve state-of-the-art performance on VIBE Bench among both open-source and closed-source models.

Our main contributions are as follows:
\begin{itemize}
    \setlength\itemsep{0pt}
    \item We perform empirical studies of scribble-guided image editing and show that instruction-level generalization is a much larger challenge than image-domain generalization, identifying the main limitation of current open-source models.
    \item We propose a unified framework for scribble-guided image editing that integrates a Coverage-then-Realism Curriculum, Multi-Task Mosaicking, and an Edit-Focused Loss to improve editing coverage, multi-task capability, and training effectiveness.
    \item Our method achieves state-of-the-art results on VIBE Bench for both single-task and multi-task scribble-guided image editing.

\end{itemize}

%% file: sections/2_related_works.tex
\section{Related Work}
\label{sec:related}

\paragraph{Text-To-Image generation.}
Recent image generation has shifted toward scalable DiT and MM-DiT backbones trained with rectified-flow or flow-matching objectives, yielding stronger prompt following, typography, high-resolution synthesis, and multimodal conditioning~\citep{peebles2023dit,lipman2023flowmatching,esser2024sd3}. These systems increasingly blur the boundary between generation and editing, supporting text-to-image synthesis, image-to-image transformation, reference conditioning, multi-image composition, and text rendering within a unified interface~\citep{wu2025qwenimage,bfl2025fluxkontext,bytedance2025seedream4,deepmind2025nanobananapro}. In parallel, controllable generation methods inject spatial conditions such as boxes, edges, depth, segmentation maps, poses, and sketches through adapters or grounding modules~\citep{zhang2023controlnet,mou2024t2iadapter,li2023gligen,zhao2023unicontrolnet}, making modern generators powerful visual backbones but also highlighting the need for lightweight spatial interaction.

\paragraph{Image-To-Image editing.}
Image editing transforms a source image under a user's instruction while preserving irrelevant content. Text-guided methods provide a convenient interface through prompt-level manipulation, inversion, or instruction tuning~\citep{hertz2022prompttoprompt,mokady2023nulltext,kawar2023imagic,brooks2023instructpix2pix,shuai2024survey}, but text-instruction alone often underspecifies the target region, object identity, or geometry. Mask-based and region-aware methods improve localization with explicit or automatically estimated masks~\citep{avrahami2021blended,couairon2022diffedit,mao2023magedit}, while spatial-control methods use sketches, edges, depth, or segmentation maps~\citep{mao2023sketchffusion,zhang2023controlnet,mou2024t2iadapter}; these signals are effective but can be tedious or too structured for casual interaction. Scribbles offer a lighter alternative: recent scribble- and sketch-guided methods show that coarse visual marks can complement language for spatially grounded edits~\citep{xia2025dreamomni3,zhang2026scribblesense,qu2025replan}. 
However, most prior work remains centered on single-scribble edits. Recent benchmarks such as VIBE~\citep{zhang2026vibe} reveal that performance degrades substantially in multi-task scribble-guided editing, and also show large performance gaps across open-source and closed-source editing models. Our work aims to systematically improve scribble-guided image editing and push an open-source model to state-of-the-art performance.

%% file: sections/3_empirical_study.tex
\section{Empirical Study: Generalization Asymmetries}
\label{sec:study}
This section studies a question in scribble-based image editing: which aspects of generalization can be inherited from pre-trained priors, and which must be learned from carefully curated training data? To answer this question, we conduct three experiments that analyze generalization along three axes: (i) individual editing tasks, (ii) image domains, and (iii) the transition from single-task to multi-task learning settings. For a controlled comparison, all experiments use the same backbone model (Qwen-Image-Edit-2511), LoRA hyperparameters, optimizer configuration, and training schedule. The only difference lies in the training data. These experiments use our editing dataset in Section \ref{sec:data:object}.

\subsection{Study 1: Instruction-Level Cross-Task Transfer Requires Full Task Coverage}
\label{sec:study:s1}

We fine-tune four expert models, each specialized for a distinct editing task—addition (AD), removal (RM), replacement (RP), and translation (TR)—and perform cross-task evaluation on the VIBE single-task test set (Table \ref{tab:cross-task}). 

The results indicate that, although Qwen-Image-Edit-2511 possesses general editing capabilities on non-scribble editing tasks, fine-tuning on data from only one scribble-editing task does not enable the model to effectively generalize these capabilities to other scribble-editing tasks. Specifically, models trained separately on each of the four editing tasks do not perform as well on unseen tasks as they do on the task for which training data are available. For example, when trained only on AD data, performance on RM and TR during testing is worse than the pre-fine-tuning baseline. Similarly, when trained only on RM or RP data, performance on the TR task is also lower than the baseline before training. Although training on TR data leads to positive improvements on the other tasks, its performance on the RM task still remains substantially below that of a model trained with RM data.


Consequently, the training data must cover all types of editing tasks, which constitutes the primary motivation for our low-cost data synthesis strategy described in Section \ref{sec:data:object}.

\begin{table}[h]
\centering
\caption{Cross-task transfer performance of scribble instructions on VIBE. Rows denote training tasks, columns denote evaluation tasks, and bold text indicates diagonal results. }
\label{tab:cross-task}
\begin{tabular}{l|cccc}
\toprule
Train \textbackslash Eval & AD & RM & RP & TR \\
\midrule
Baseline & $86.73$ & $61.15$ & $74.73$ & $30.42$ \\
Only AD & \textbf{86.50} & 37.75 & 80.21 & 9.43 \\
Only RM & 87.62 & \textbf{97.18} & 85.50 & 16.99 \\
Only RP & 87.93 & 97.21 & \textbf{81.51} & 21.14 \\
Only TR & 88.84 & 82.54 & 80.09 & \textbf{84.01} \\
\bottomrule
\end{tabular}
\end{table}

\subsection{Study 2: Image-Domain Generalization Has Limited Impact on Scribble-Guided Editing}
\label{sec:study:s2}


Beyond cross-task transfer, we further examine synthetic-to-real cross-domain generalization while keeping the edit tasks fixed. As described in Section~\ref{sec:data:object}, we construct both synthetic and real-image datasets and train separate models on equal 4.5K subsets from each domain. Both models are evaluated exclusively on VIBE's real-world split, and the reported scores are averaged over all tasks. This comparison therefore isolates the effect of replacing in-domain real-image training data with cross-domain synthetic data. Table~\ref{tab:cross-domain} shows that the synthetic-trained model nearly matches the real-trained model in Instruction Adherence (85.62 vs.\ 86.27), with only a 0.65-point drop. The effect is larger for Visual Coherence, a metric more directly tied to low-level realism and local image integration, where synthetic training trails real-image training by 4.16 points (87.75 vs.\ 91.91).

\begin{table}[h]
\centering
\caption{VIBE real-world evaluation of models trained on in-domain real-world data versus cross-domain synthetic data.}
\label{tab:cross-domain}
\begin{tabular}{lcc}
\toprule
Training Data & Instruction Adherence & Visual Coherence \\
\midrule
Real-World (In-Domain) & 86.27 & 91.91 \\
Synthetic (Cross-Domain) & 85.62 & 87.75 \\
\bottomrule
\end{tabular}
\end{table}

These results indicate that the synthetic-to-real domain gap affects visual realism more than instruction following. Although synthetic renderings and real-world images differ substantially in illumination, depth of field and texture statistics, the model trained only on synthetic images still learns to follow scribble-guided editing instructions on real images. Its main limitation is visual integration: edited regions may show artifacts because low-level properties such as luminance and texture are not always matched to the natural background. Compared with the much larger cross-task failures observed in Section~\ref{sec:study:s1}, the cross-domain gap appears relatively easy to bridge, suggesting that broad task coverage is more important than exhaustive image-domain coverage.

\subsection{Study 3: Multi-Task Generalization Requires Higher-Order Paired Supervision}
\label{sec:study:s3}

A stratified analysis over the number of tasks in VIBE shows that existing image editing models perform much worse on multi-task samples than on single-task samples. We reproduce this in a controlled setting: a model trained on a heterogeneous set of single-task examples can perform well on individual tasks but still exhibits severe logical errors on multi-task inputs (a single image with $k$ scribbles and $k$ matching instructions, $k>1$). The main failures are (i) \emph{instruction crosstalk}, where an edit for one scribble is applied to the wrong region, and (ii) \emph{instruction omission}, where some requested edits are ignored. A natural hypothesis is that this degradation stems from out-of-distribution (OOD) visual statistics of “single-image multi-scribble” inputs. We design a targeted falsification experiment to test this.

\begin{table}[h]
\centering
\caption{VIBE performance comparison between single-task and multi-task training with visual distractor.}
\label{tab:falsification}
\begin{tabular}{lcc}
\toprule
Training Data & 2 Tasks & 3 Tasks \\
\midrule
Single-task data only & 75.07 $\pm$ 0.74 & 64.63 $\pm$ 2.57 \\
With distractor scribbles & 75.72 $\pm$ 1.66 & 62.85 $\pm$ 7.76 \\
\bottomrule
\end{tabular}
\end{table}

We build a single-task training image set with visual distractor: each image has exactly one valid “scribble–instruction–edit” tuple, plus 1–3 additional distractive scribbles without instructions, randomly overlaid. Supervision remains strictly single-task, but the input image distribution now matches that of multi-task evaluation. If OOD visual statistics were the main issue, this distractor scheme should noticeably reduce multi-task failures.

The results do not support this (Table \ref{tab:falsification}): the model improves by only +0.65 on 2-Task and degrades by -1.78 on 3-Task, both within noise. Thus, even though distractor examples match the multi-scribble visuals, limiting supervision to a single tuple per image prevents the model from learning the full joint mapping over “multi-scribble, multi-instruction, multi-edit” configurations.

Therefore, the core cause of multi-task editing failure is the lack of supervision on complete higher-order tuples $\langle \text{multi-scribble image}, \text{multi-instruction prompt}, \text{multi-edit ground truth} \rangle$. Robust instance-level bindings $\text{instruction}_i \mapsto \text{scribble}_i \mapsto \text{edit}_i$ emerge only when jointly paired data are explicitly provided.

\subsection{Implications: Motivation for Our Methods}
\label{sec:study:conclusion}

The three probe studies clarify how different editing capabilities should be supervised. Studies 1 and 3 show that scribble-guided editing does not reliably generalize to new task settings: models fail to transfer consistently across editing operations, and single-task supervision does not yield robust multi-task editing even when the visual input distribution is matched. These capabilities therefore require explicit coverage in the training data. By contrast, Study 2 shows that instruction following transfers well from synthetic to real images, with the main domain gap lying in visual realism.

The key insight is that the image-domain gap mainly affects visual realism, whereas instruction logic must be learned from explicit supervision. This has two implications. First, it supports synthesizing multi-task tuples by mosaicking low-cost single-task samples: even if mosaicked images deviate from natural image distributions, the instruction bindings they provide still transfer to real images. Second, it suggests a progressive data construction and training strategy. Accordingly, our training strategy first uses a large-scale, low-cost synthetic dataset (Section \ref{sec:data}) to learn editing intents and multi-task bindings, and then uses a small set of real-world images to address the remaining visual-realism gap.

%% file: sections/4_methods.tex
\section{Methods}
\label{sec:data}

\subsection{Problem Formulation}
\label{sec:data:setup}

Given an input image $I$ overlaid with user scribbles $S$ and a textual instruction $T$, the model produces an edited image $\tilde{I}$ that satisfies $T$ within the region(s) marked by $S$ while keeping the rest of the image unchanged. We render $S$ directly into the input image so that the interface $(I \oplus S, T) \to \tilde{I}$ matches that of standard image editing models~\citep{wu2025qwenimage}, requiring no architectural change.

\begin{figure}[t]
  \centering
  \includegraphics[width=1.0\linewidth]{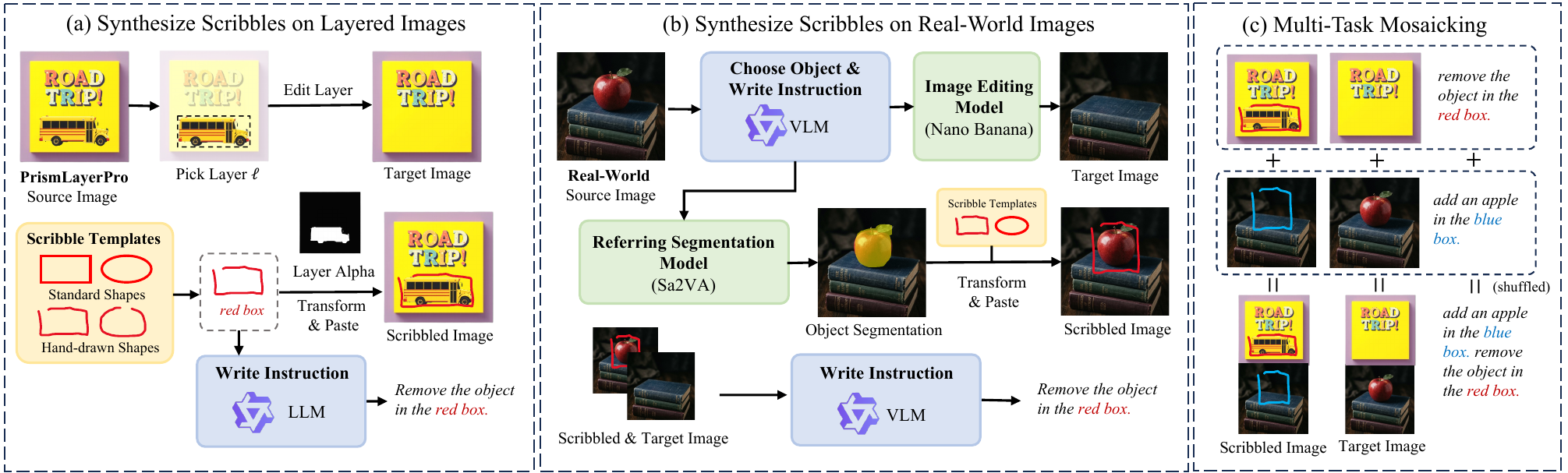}
  \caption{Overview of our data construction pipelines: (a) synthesize scribbles on layered images~\citep{chen2025prismlayers}; (b) synthesize scribbles on real-world images by image editing and referring segmentation. (c) multi-task training samples are produced by mosaicking single-task samples.}
  \label{fig:pipeline}
\end{figure}

\subsection{Coverage-then-Realism Curriculum Data Pipeline}
\label{sec:data:object}

Guided by the generalization findings in Section~\ref{sec:study}, we decompose data construction into two stages: Stage 1, a large-scale synthetic pool that provides broad task 
coverage with exact supervision, and Stage 2, a small set of curated real-world samples that inject visual realism. 
Figure~\ref{fig:pipeline} summarizes the overall pipeline.

\paragraph{Stage 1: Synthetic Scribbles on Layered Images.}
To construct Stage 1 at scale, we need an image source where single-object edits can be generated without per-example masks or learned inpainting. PrismLayersPro~\citep{chen2025prismlayers} provides such a source: its multi-layer RGBA compositions allow each single-object edit to be implemented by re-rendering the underlying layer stack. As illustrated in Figure~\ref{fig:pipeline} (a), our pipeline proceeds in three steps. \textbf{(1) Layer-level editing.} Given a source image $I$, we select a target layer $\ell$ and apply the task-specific edit to obtain $\tilde{I}$. \textbf{(2) Scribble synthesis.} We sample a template from a library that mixes geometric primitives (rectangles, ellipses) with hand-drawn presets, transform it with the bounding box or alpha-matte contour of $\ell$, and randomize its color, thickness, and placement to produce a scribble on $I$. \textbf{(3) Instruction generation.} An LLM is prompted with the task type, scribble type, and scribble color to write the matching instruction. This design scales effortlessly to a large synthetic dataset.

\paragraph{Stage 2: Curated Scribbles on Real-World Scenes.}
Stage 1 alone leaves a visible gap on real-world images: composited layers do not match the lighting, perspective, or background blending of real-world scenes. Stage 2 closes this gap with a small set of real-world images. As shown in Figure~\ref{fig:pipeline} (b), given a real photograph $I$ and a target task, we first ask a VLM to propose a candidate object and write a coarse editing instruction $T$. We then use an image editing model (e.g., Nano Banana Pro~\citep{deepmind2025nanobananapro}) to produce $\tilde{I}$ from $(I, T)$, and a referring segmentation model (e.g., Sa2VA~\citep{yuan2025sa2va}) to localize the affected region, around which we synthesize a scribble using the same templates library as in Stage 1. After that, we prompt a VLM with the scribbled image $I \oplus S$ and the edited target $\tilde{I}$ to write the final scribble-aware instruction. Due to the occasional failure of image editing, we filter the candidates by human review: from $\sim\!20\textnormal{K}$ raw candidates we keep $3.8\textnormal{K}$ samples and add $700$ samples with human-drawn scribbles, yielding $4.5\textnormal{K}$ pairs for Stage-2 fine-tuning.

\paragraph{Multi-Task Mosaicking.}
\label{sec:data:mt}

The falsification experiment in Section~\ref{sec:study:s3} attributes the multi-task performance gap to insufficient supervision on \emph{complete} higher-order tuples; therefore, our data construction must generate such tuples at scale. However, generating these samples from scratch is difficult: the individual edits must not interfere with each other, which is hard to guarantee with either pipeline above. We therefore propose \emph{Multi-Task Mosaicking} as a simple yet effective alternative. As illustrated in Figure~\ref{fig:pipeline} (c), given $k\!\in\!\{2,4\}$ single-task samples, we mosaic their inputs into a $1\!\times\!2$, $2\!\times\!1$, or $2\!\times\!2$ grid, arrange the corresponding targets in the same layout, and concatenate the $k$ instructions in a random order. The result is a composite sample carrying $k$ spatially separated scribble--instruction pairs in the standard format. To eliminate binding ambiguity, the $k$ sources are sampled with mutually distinct scribble colors. This construction is supported by the generalization findings in Sections~\ref{sec:study:s2} and~\ref{sec:study:s3}: although mosaicked images are visibly off-distribution, Study 2 shows that such image-domain shifts mainly affect visual realism rather than instruction following.

\subsection{Edit-Focused Loss}
\label{sec:training:roi}

In scribble-guided editing, only the marked region should change while most pixels should remain unchanged. A standard pixel-uniform flow-matching loss assigns equal weight to all pixels, so optimization is dominated by the unchanged background rather than the edited region. Stage-1 synthetic data allow us to correct this imbalance: because each edit is generated by an exact layer operation, we can directly obtain a pixel-level mask of the edited region. We therefore up-weight the loss inside this mask.

Concretely, for Stage-1 samples we obtain the binary region mask $M \in \{0,1\}^{H \times W}$ by thresholding the per-pixel difference $|I - \tilde{I}|$ between source and target images and down-sampling it to the latent resolution. Let $v_\theta(x_t, t, c)$ and $v^\ast$ denote the predicted and target velocities under the standard flow-matching objective, and define the whole-image and mask-area-normalized edit losses
\begin{align}
    \mathcal{L}_{\mathrm{whole}} &\;=\; \mathbb{E}\!\left[\, \big\| v_\theta(x_t, t, c) - v^\ast \big\|^{2} \,\right], \\
    \mathcal{L}_{\mathrm{edit}}   &\;=\; \mathbb{E}\!\left[\, M \odot \big\| v_\theta(x_t, t, c) - v^\ast \big\|^{2} \,\right],
\end{align}
where $\odot$ denotes element-wise multiplication. We train under the additive objective
\begin{equation}
    \mathcal{L} \;=\; \mathcal{L}_{\mathrm{whole}} \;+\; \lambda\, \mathcal{L}_{\mathrm{edit}},
    \label{eq:roi}
\end{equation}
with $\lambda = 0.1$. The value of $\lambda$ is chosen so that the two loss terms are of similar magnitudes. We use this edit-focused term only during Stage-1 training, as exact edit-region masks are not available in Stage 2: edits on real images often induce diffuse changes in shadows, illumination, and local blending, making the effective edited region ambiguous.

%% file: sections/5_experiments.tex
\section{Experiments}
\label{sec:experiments}

\subsection{Training Details}
\label{sec:training:backbone}

We adopt Qwen-Image-Edit-2511~\citep{wu2025qwenimage}, an open-source image editing model. To minimize both computational overhead and architectural modifications, we paste the scribble $S$ directly onto the input image (Section~\ref{sec:data:setup}) rather than introducing a separate conditioning branch. This design preserves the backbone's native image-text interface, ensuring that our training recipe can be seamlessly transferred to any standard image editing model. For optimization, we fine-tune the model using a standard LoRA adapter~\citep{hu2021lora} with a rank of $32$ and a learning rate of $10^{-4}$. Stage 1: Pre-training on synthetic data with Edit-Focused Loss and Multi-Task Mosaicking. We train the model on the full synthetic dataset comprising $180\textnormal{K}$ samples (Section~\ref{sec:data:object}), augmented via Multi-Task Mosaicking (Section~\ref{sec:data:mt}). Single-task and mosaicked multi-task samples are mixed at a $4{:}1$ ratio. The model is optimized using the objective defined in Equation~\ref{eq:roi}. Stage 2: Fine-tuning on real-world data. We then fine-tune the Stage-1 model on the $4.5\textnormal{K}$ curated real-world samples of Section~\ref{sec:data:object}. 

\subsection{Evaluation Protocol}
\label{sec:experiments:setup}

We evaluate scribble-guided object editing using the Deictic Level of the VIBE benchmark~\citep{zhang2026vibe}. The evaluation covers four single-task operations: Addition (AD), Removal (RM), Replacement (RP), and Translation (TR), as well as multi-task settings involving 2 Tasks and 3 Tasks. Following the standard VIBE protocol, we employ its default LMM-as-a-judge evaluator (GPT-5.1) and report the average score over three independent runs. We compare our final model against leading closed-source models (Nano Banana Pro~\citep{deepmind2025nanobananapro}, Qwen-Image-2.0 Pro~\citep{wu2025qwenimage}, Wan~2.7 Pro~\citep{mao2026wan}, Seedream~4.5~\citep{bytedance2025seedream4}, and Seedream~5.0~\citep{seedream5_2025}) and open-source models (the unmodified Qwen-Image-Edit-2511 baseline~\citep{wu2025qwenimage}, Qwen-Image-Edit-2509~\citep{wu2025qwenimage}, and FLUX2-dev~\citep{flux2_2025}).

\subsection{Quantitative Results}
\label{sec:experiments:main}

Table~\ref{tab:main-object} reports the VIBE Deictic-Level scores. Our model demonstrates substantial improvements over the unmodified Qwen-Image-Edit-2511 baseline across all tasks. It obtains the best 1-Task average and the best scores on Addition and Translation, while remaining competitive on Removal and Replacement. On multi-task evaluations, it ranks second on both 2 Tasks and 3 Tasks, narrowly behind Seedream~5.0 and ahead of Nano Banana Pro.

\begin{table}[h]
  \caption{VIBE Deictic-Level scores on single-task (AD, RM, RP, TR; 1 Task reports their average) and multi-task (2 Tasks, 3 Tasks) settings. Best per column in bold and second-best underlined.}
  \label{tab:main-object}
  \centering
  \small
  \setlength{\tabcolsep}{4pt}
  \begin{tabular}{l|cccc|c|cc}
    \toprule
    {Method} & AD & RM & RP & TR & 1 Task & 2 Tasks & 3 Tasks \\
    \midrule
    \multicolumn{8}{l}{\emph{Closed-source}} \\
    Nano Banana Pro                                           & $82.17$ & $94.07$ & $\underline{88.26}$ & $74.80$ & $84.83$ & $80.22$ & $75.48$ \\
    Qwen-Image-2.0 Pro                                        & $45.00$ & $60.55$ & $39.12$ & $36.37$ & $45.26$ & $35.31$ & $43.26$ \\
    Wan 2.7 Pro                                               & $90.51$ & $93.92$ & $\mathbf{88.34}$ & $51.47$ & $81.06$ & $69.01$ & $68.55$ \\
    Seedream 4.5                                              & $83.31$ & $\mathbf{95.26}$ & $84.00$ & $48.03$ & $77.65$ & $70.20$ & $61.28$ \\
    Seedream 5.0                                              & $\underline{90.58}$ & $91.88$ & $87.16$ & $\underline{81.56}$ & $\underline{87.80}$ & $\mathbf{84.39}$ & $\mathbf{77.45}$ \\
    \midrule
    \multicolumn{8}{l}{\emph{Open-source}} \\
    Qwen-Image-Edit-2511 (baseline)                           & $86.73$ & $61.15$ & $74.73$ & $30.42$ & $63.26$ & $57.90$ & $59.05$ \\
    Qwen-Image-Edit-2509                                      & $55.28$ & $14.38$ & $30.13$ & $14.48$ & $28.57$ & $-$     & $-$     \\
    FLUX2-dev                                                 & $64.57$ & $8.00$  & $54.40$ & $5.58$  & $33.14$ & $-$     & $-$     \\
    Ours                                                      & $\mathbf{90.68}$ & $\underline{94.79}$ & $86.53$ & $\mathbf{84.91}$ & $\mathbf{89.23}$ & $\underline{83.27}$ & $\underline{76.68}$ \\
    \bottomrule
  \end{tabular}
\end{table}

\subsection{Qualitative Results}
\label{sec:experiments:qualitative}

\begin{figure}[t]
  \centering
  \includegraphics[width=0.9\linewidth]{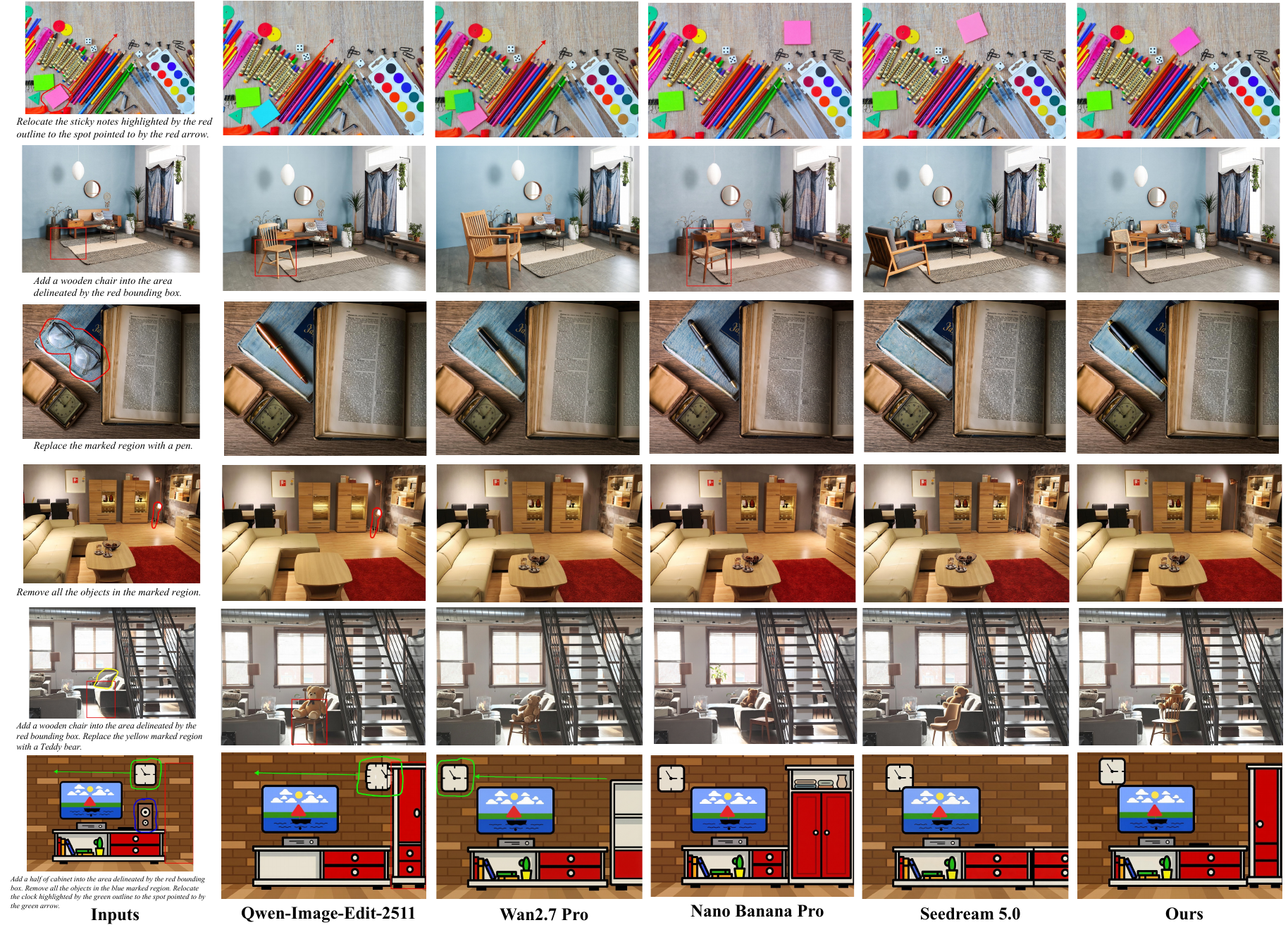}
  \caption{Qualitative comparison across single-task and multi-task editing.}
  \label{fig:qualitative}
\end{figure}

Figure~\ref{fig:qualitative} shows side-by-side comparisons across object single-task and multi-task editing. Two common failure patterns emerge among the strongest baselines. First, in single-task editing, existing models occasionally misinterpret instructions, modify unintended objects, generate inconsistent appearances, or struggle to maintain visual realism (e.g., leaving lingering shadows). Second, in multi-task editing, these models frequently omit certain editing tasks. In contrast, our method consistently produces faithful edits across both settings with a significantly lower failure rate.

\subsection{Component Ablations}
\label{sec:experiments:ablation}

The Stage-1 variants presented in Table~\ref{tab:stage1-ablation} allow us to isolate the contribution of each training component.

\begin{table}[h]
  \caption{Stage-1 component ablations on VIBE Deictic-Level scores. Best per column in bold.}
  \label{tab:stage1-ablation}
  \centering
  \small
  \setlength{\tabcolsep}{4pt}
  \begin{tabular}{l|cccc|c|cc}
    \toprule
    {Method} & AD & RM & RP & TR & 1 Task & 2 Tasks & 3 Tasks \\
    \midrule
    Stage 1 w/o Edit-Focused Loss
      & $88.37$ & $95.84$ & $83.82$ & $81.59$ & $87.41$ & $81.90$ & $71.72$ \\
    Stage 1 w/o Mosaicking
      & $\mathbf{90.26}$ & $\mathbf{98.29}$ & $\mathbf{85.43}$ & $\mathbf{83.52}$ & $\mathbf{89.38}$ & $73.21$ & $52.18$ \\
    Stage 1
      & $88.39$ & $98.18$ & $84.60$ & $81.24$ & $88.10$ & $\mathbf{82.65}$ & $\mathbf{73.29}$ \\
    \bottomrule
  \end{tabular}
\end{table}

\paragraph{Edit-Focused Loss provides consistent refinements.}
Removing the Edit-Focused Loss (\textit{Stage 1 w/o Edit-Focused Loss} vs.\ \textit{Stage 1}) results in small drops on most VIBE tasks, especially on Removal, Replacement, and 2 Tasks. The magnitude of this effect aligns with our rationale in Section~\ref{sec:training:roi}: since the Edit-Focused Loss acts as a straightforward importance-sampling mechanism based on a spatial prior, it is expected to yield consistent refinements rather than dramatic performance leaps.

\begin{figure}[h]
  \centering
  \includegraphics[width=0.7\linewidth]{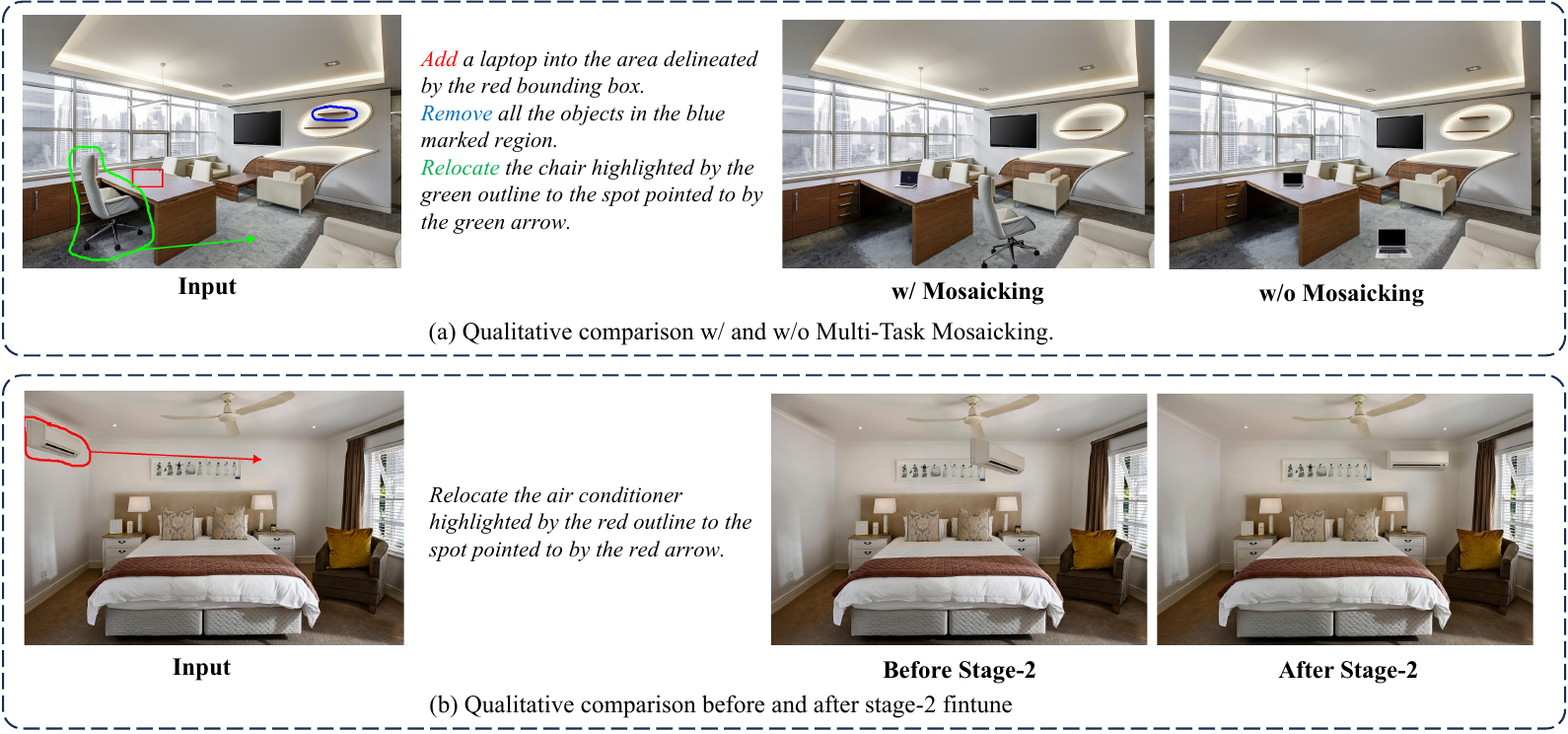}
  \caption{Qualitative comparisons for the two training components. (a) Multi-scribble editing with and without Multi-Task Mosaicking. (b) Editing realism before and after the Stage-2 fine-tune.}
  \label{fig:real-and-mt-comparison}
\end{figure}

\paragraph{Multi-Task Mosaicking closes the multi-task gap.}
Removing the Multi-Task Mosaicking (\textit{Stage 1 w/o Mosaicking} vs.\ \textit{Stage 1}) causes a severe performance drop on the 2 Tasks and 3 Tasks settings, while leaving the 1 Task score largely unaffected (and even slightly improved, as the model no longer needs to accommodate visually unfamiliar mosaicked images). Figure~\ref{fig:real-and-mt-comparison} (a) shows the same pattern qualitatively: without mosaicked data, the model often ignores one instruction or binds an edit to the wrong scribble, whereas Stage 1 applies each instruction to its intended target. This pattern matches the prediction made by Study~3 (Section~\ref{sec:study:s3}): the multi-task performance gap stems primarily from a lack of paired complex supervision rather than visibly off-distribution. By supplying these paired multi-task tuples through mosaicking, the gap is effectively closed.

\paragraph{Stage-2 fine-tuning enhances realism without compromising task coverage.}
Applying the Stage-2 fine-tuning (\textit{Stage 1 + Stage 2} vs.\ \textit{Stage 1}) yields modest gains in the overall VIBE score but provides a substantial boost to the Visual Coherence ($\mathcal{VC}$) sub-score (detailed in Table~\ref{tab:stage2-vc}). This specific improvement directly addresses the vulnerability identified by Study 2 (Section~\ref{sec:study:s2}), which flagged visual realism as the primary weakness of a purely synthetic-trained model. Figure~\ref{fig:real-and-mt-comparison} (b) qualitatively illustrates this effect: while the Stage-1 model successfully executes the editing instructions, the Stage-2 model achieves significantly more natural perspective and object-background blending. Crucially, these results quantitatively confirm our hypothesis: domain realism is a separable attribute that can be effectively patched with a small amount of real-world data, without compromising the broad task coverage established in Stage 1.

\begin{table}[h]
  \caption{Visual Coherence ($\mathcal{VC}$) sub-scores on VIBE before and after the Stage-2 real-world fine-tune.}
  \label{tab:stage2-vc}
  \centering
  \small
  \setlength{\tabcolsep}{12pt}
  \begin{tabular}{l|ccc}
    \toprule
    {Method} & 1 Task & 2 Tasks & 3 Tasks \\
    \midrule
    Stage 1                       & $89.61$ & $84.20$ & $77.22$ \\
    Stage 1 + Stage 2 (final)     & $\mathbf{93.81}$ & $\mathbf{88.40}$ & $\mathbf{82.96}$ \\
    \bottomrule
  \end{tabular}
\end{table}

%% file: sections/6_conclusion.tex
\section{Conclusion}
\label{sec:conclusion}


In this paper, we investigated scribble-guided image editing and identified a clear asymmetry in model generalization: instruction-level generalization is significantly more challenging than image-domain generalization. To address this bottleneck, we proposed three simple yet effective strategies: a Coverage-then-Realism Curriculum to balance broad instruction learning with generation realism, Multi-Task Mosaicking to construct multi-task training samples at nearly zero cost, and an Edit-Focused Loss to enhance learning efficiency and precision. Together, these strategies enable our model to achieve state-of-the-art performance on the VIBE Bench for both single- and multi-task editing. We will publicly release our dataset and model to support future research.

%% file: sections/X_supply.tex
\section{Full Experimental Results}
\label{sec:supp:full-results}

Table~\ref{tab:supp-main-object-full} reports the full VIBE Deictic-Level results corresponding to Table~\ref{tab:main-object} in the main paper, including the mean and standard deviation over three independent runs when available. Results for Nano Banana Pro, Qwen-Image-Edit-2509, and FLUX2-dev are taken directly from the original VIBE paper. For these methods, entries are left blank when the corresponding score or standard deviation is not reported in the original paper.

\begin{table}[h]
  \caption{Full VIBE Deictic-Level scores with mean and standard deviation over 3 independent runs.}
  \label{tab:supp-main-object-full}
  \centering
  \scriptsize
  \setlength{\tabcolsep}{3pt}
  \resizebox{\linewidth}{!}{
  \begin{tabular}{l|cccc|c|cc}
    \toprule
    {Method} & AD & RM & RP & TR & 1 Task & 2 Tasks & 3 Tasks \\
    \midrule
    \multicolumn{8}{l}{\emph{Closed-source}} \\
    Nano Banana Pro
      & $82.17 \pm 0.66$ & $94.07 \pm 0.73$ & $88.26 \pm 1.97$ & $74.80 \pm 2.13$ & $84.83$ & $80.22$ & $75.48$ \\
    Qwen-Image-2.0 Pro
      & $45.00 \pm 1.94$ & $60.55 \pm 4.02$ & $39.12 \pm 1.10$ & $36.37 \pm 1.89$ & $45.26 \pm 2.48$ & $35.31 \pm 1.09$ & $43.26 \pm 6.52$ \\
    Wan 2.7 Pro
      & $90.51 \pm 1.13$ & $93.92 \pm 1.57$ & $88.34 \pm 0.75$ & $51.47 \pm 0.57$ & $81.06 \pm 1.07$ & $69.01 \pm 2.76$ & $68.55 \pm 6.23$ \\
    Seedream 4.5
      & $83.31 \pm 0.82$ & $95.26 \pm 0.51$ & $84.00 \pm 2.70$ & $48.03 \pm 1.59$ & $77.65 \pm 1.64$ & $70.20 \pm 3.07$ & $61.28 \pm 1.63$ \\
    Seedream 5.0
      & $90.58 \pm 1.23$ & $91.88 \pm 1.15$ & $87.16 \pm 0.52$ & $81.56 \pm 1.68$ & $87.80 \pm 1.22$ & $84.39 \pm 2.21$ & $77.45 \pm 2.36$ \\
    \midrule
    \multicolumn{8}{l}{\emph{Open-source}} \\
    Qwen-Image-Edit-2511
      & $86.73 \pm 0.25$ & $61.15 \pm 3.58$ & $74.73 \pm 0.37$ & $30.42 \pm 4.00$ & $63.26 \pm 2.69$ & $57.90 \pm 2.60$ & $59.05 \pm 4.62$ \\
    Qwen-Image-Edit-2509
      & $55.28 \pm 2.40$ & $14.38 \pm 3.05$ & $30.13 \pm 0.92$ & $14.48 \pm 0.57$ & $28.57$ & -- & -- \\
    FLUX2-dev
      & $64.57 \pm 1.18$ & $8.00 \pm 1.73$ & $54.40 \pm 1.71$ & $5.58 \pm 1.09$ & $33.14$ & -- & -- \\
    \midrule
    \multicolumn{8}{l}{\emph{Ours}} \\
    Stage 1 w/o Edit-Focused Loss
      & $88.37 \pm 0.92$ & $95.84 \pm 0.44$ & $83.82 \pm 0.67$ & $81.59 \pm 0.33$ & $87.41 \pm 0.63$ & $81.90 \pm 2.80$ & $71.72 \pm 5.73$ \\
    Stage 1 w/o Mosaicking
      & $90.26 \pm 1.85$ & $98.29 \pm 1.12$ & $85.43 \pm 1.10$ & $83.52 \pm 1.25$ & $89.38 \pm 1.36$ & $73.21 \pm 2.56$ & $52.18 \pm 6.82$ \\
    Stage 1
      & $88.39 \pm 0.48$ & $98.18 \pm 1.02$ & $84.60 \pm 0.10$ & $81.24 \pm 2.80$ & $88.10 \pm 1.51$ & $82.65 \pm 2.25$ & $73.29 \pm 3.48$ \\
    Stage 1 + Stage 2 (final)
      & $90.68 \pm 1.00$ & $94.79 \pm 0.85$ & $86.53 \pm 1.41$ & $84.91 \pm 1.52$ & $89.23 \pm 1.23$ & $83.27 \pm 4.28$ & $76.68 \pm 2.42$ \\
    \bottomrule
  \end{tabular}
  }
\end{table}

\section{Compute Resources Details}
\label{sec:supp:compute}

For the main training runs, we fine-tune the model on $16$ NVIDIA H20 GPUs, each with $96$GB of device memory. We use a total batch size of $64$ across all GPUs. The two-stage training schedule consists of $4{,}000$ optimization steps for Stage~1 and $500$ optimization steps for Stage~2. In our implementation, training requires approximately $43$ minutes per $100$ optimization steps on this hardware configuration. This corresponds to approximately $32.3$ wall-clock hours in total for one complete Stage~1+Stage~2 training run.

\section{More Qualitative Comparisons}
\label{sec:supp:more-qualitative}

\begin{figure}[p]
  \centering
  \includegraphics[width=1.0\linewidth]{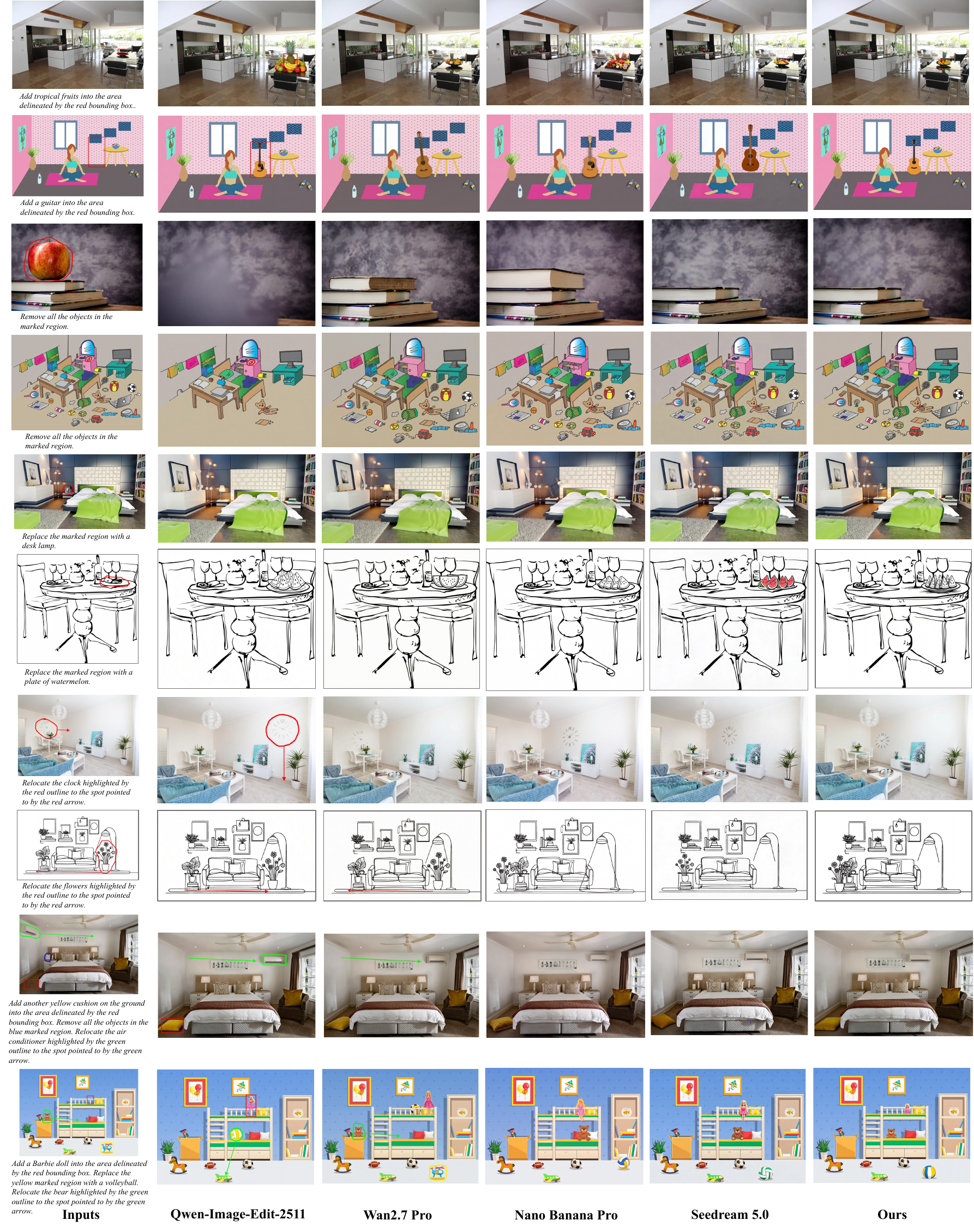}
  \caption{More qualitative results for scribble-guided editing.}
  \label{fig:supp-more-qualitative}
\end{figure}

Figure~\ref{fig:supp-more-qualitative} provides more qualitative comparisons.

\section{Scribble-Guided Text Editing as an Application}
\label{sec:supp:text}

Beyond object editing in the main paper, we explored scribble-guided \emph{text} editing as a downstream application of the same recipe. The text-editing data construction, the VIBE-Text benchmark, and current results are presented here as a self-contained appendix. We keep text editing outside the main training recipe to preserve a clean separation between the paper's central object-editing claim and a distinct application that requires text-specific supervision, re-layout behavior, and evaluation criteria.

\subsection{Text Editing Dataset}
\label{sec:supp:text:data}

Text editing is not a special case of object editing: insertion, deletion, and replacement can change the length of a line and force the model to \emph{re-layout} the surrounding text, which pure object-editing supervision provides no signal for. We build a dedicated text-editing dataset and keep it fully synthetic because rendered text gives exact pixel-level ground truth, while real paired before/after text-editing examples with localized scribbles are difficult to collect at scale.

We cover four tasks: \textbf{Insertion}, \textbf{Deletion}, \textbf{Text Replacement}, and \textbf{Style Modification}. The first three operate on character content and always require re-layout; Style Modification triggers a re-layout only when the modified attribute changes character footprint (e.g., font size or spacing) and leaves the layout untouched otherwise (e.g., color or shadow). The instruction carries an explicit re-layout cue whenever a re-layout is needed, so the model triggers re-layout only when the prompt asks for it. To keep re-layout controllable, we restrict its scope to the last line of the edited block, leaving the rest of the page pixel-identical to the source.

The pipeline mirrors the object-editing Stage-1 of Section \ref{sec:data:object}, with rendered text in place of layered images. We first render a paragraph of mixed Chinese and English text on a solid-color canvas with randomized attributes. We then apply the edit at the source-string level and re-render with the same typography to obtain $\tilde{I}$, so that only the edited span changes pixels. The bounding box of the edited span drives scribble synthesis with randomized color, and an LLM is finally prompted with the task type, the edited text, and the scribble shape and color to write the matching instruction. This pipeline produces a synthetic text-editing dataset of $\sim\!180\textnormal{K}$ samples.

\subsection{VIBE-Text Benchmark}
\label{sec:supp:text:vibe}

VIBE's Deictic Level~\citep{zhang2026vibe} evaluates scribble-guided object editing but does not cover scribble-guided text editing. We extend it minimally so that VIBE-Text scores remain directly comparable to VIBE Deictic-Level scores. The benchmark mirrors the four text-editing tasks of Section \ref{sec:supp:text:data} and adopts VIBE's three task-level criteria with text-specific adjustments: \textbf{Instruction Adherence} ($\mathcal{IA}$) gains a binary \textit{Re-layout Compliance} sub-criterion that checks whether the model re-layouts exactly as the instruction requires; \textbf{Contextual Preservation} ($\mathcal{CP}$) requires character-level fidelity on unchanged spans while explicitly excluding instruction-required re-layouts from the violation count; and \textbf{Visual Coherence} ($\mathcal{VC}$) is augmented with text-style consistency (font, weight, size, color) and overall layout harmony. The geometric-mean aggregation and the gating of $\mathcal{VC}$ by $\mathcal{IA}$ are inherited from VIBE.

\input{sections/X_vibe_text_prompts}

\subsection{Preliminary Results on VIBE-Text}
\label{sec:supp:text:results}

We benchmark a model fine-tuned on the recipe of Section \ref{sec:data} extended with the text editing dataset of Section \ref{sec:supp:text:data}. We use Gemini-3-Flash as the VLM judge because it is more accurate on our text-specific criteria. Table~\ref{tab:supp-vibe-text} reports VIBE-Text scores across the four text-editing tasks. Our model averages $53.46$, against $44.33$ for Nano Banana Pro and $14.64$ for the unmodified Qwen-Image-Edit-2511 backbone---a $+9.13$ improvement over the closest closed-source competitor. The lead is consistent across all four tasks and is largest on Style Modification ($+13.32$ over Nano Banana Pro), suggesting that the visual constraints unique to text editing (font, color, layout) are where dedicated supervision matters most.

\begin{table}[h]
    \caption{VIBE-Text scores with mean and standard deviation across four text-editing tasks. Best per column in bold.}
    \label{tab:supp-vibe-text}
    \centering
    \small
    \setlength{\tabcolsep}{5pt}
    \resizebox{\linewidth}{!}{
    \begin{tabular}{l|cccc|c}
        \toprule
        {Method} & Deletion & Insertion & Style Mod. & Replacement & Avg. \\
        \midrule
        \multicolumn{6}{l}{\emph{Closed-source}} \\
        Nano Banana Pro    & $45.97 \pm 3.42$ & $47.47 \pm 3.50$ & $37.91 \pm 2.39$ & $45.95 \pm 1.24$ & $44.33 \pm 2.79$ \\
        Seedream 5.0       & $29.56 \pm 5.39$ & $43.80 \pm 1.33$ & $36.72 \pm 4.64$ & $42.69 \pm 2.38$ & $38.19 \pm 3.81$ \\
        Seedream 4.5       & $35.41 \pm 2.10$ & $26.58 \pm 2.50$ & $22.89 \pm 2.48$ & $43.87 \pm 2.63$ & $32.19 \pm 2.43$ \\
        Wan 2.7            & $19.27 \pm 0.42$ & $34.31 \pm 3.77$ & $34.52 \pm 1.63$ & $38.12 \pm 2.95$ & $31.55 \pm 2.54$ \\
        Wan 2.6            & $44.98 \pm 3.32$ & $29.17 \pm 2.48$ & $34.75 \pm 2.78$ & $46.05 \pm 0.96$ & $38.74 \pm 2.54$ \\
        \midrule
        \multicolumn{6}{l}{\emph{Open-source}} \\
        Qwen-Image-Edit-2511   & $\phantom{0}5.98 \pm 0.51$ & $21.88 \pm 2.19$ & $\phantom{0}8.43 \pm 2.56$ & $22.26 \pm 2.42$ & $14.64 \pm 2.09$ \\
        \midrule
        \textbf{Ours}          & $\mathbf{51.19 \pm 6.23}$ & $\mathbf{53.46 \pm 0.70}$ & $\mathbf{51.23 \pm 2.48}$ & $\mathbf{57.95 \pm 3.61}$ & $\mathbf{53.46 \pm 3.82}$ \\
        \bottomrule
    \end{tabular}
    }
\end{table}

\begin{figure}[h]
  \centering
  \includegraphics[width=1.0\linewidth]{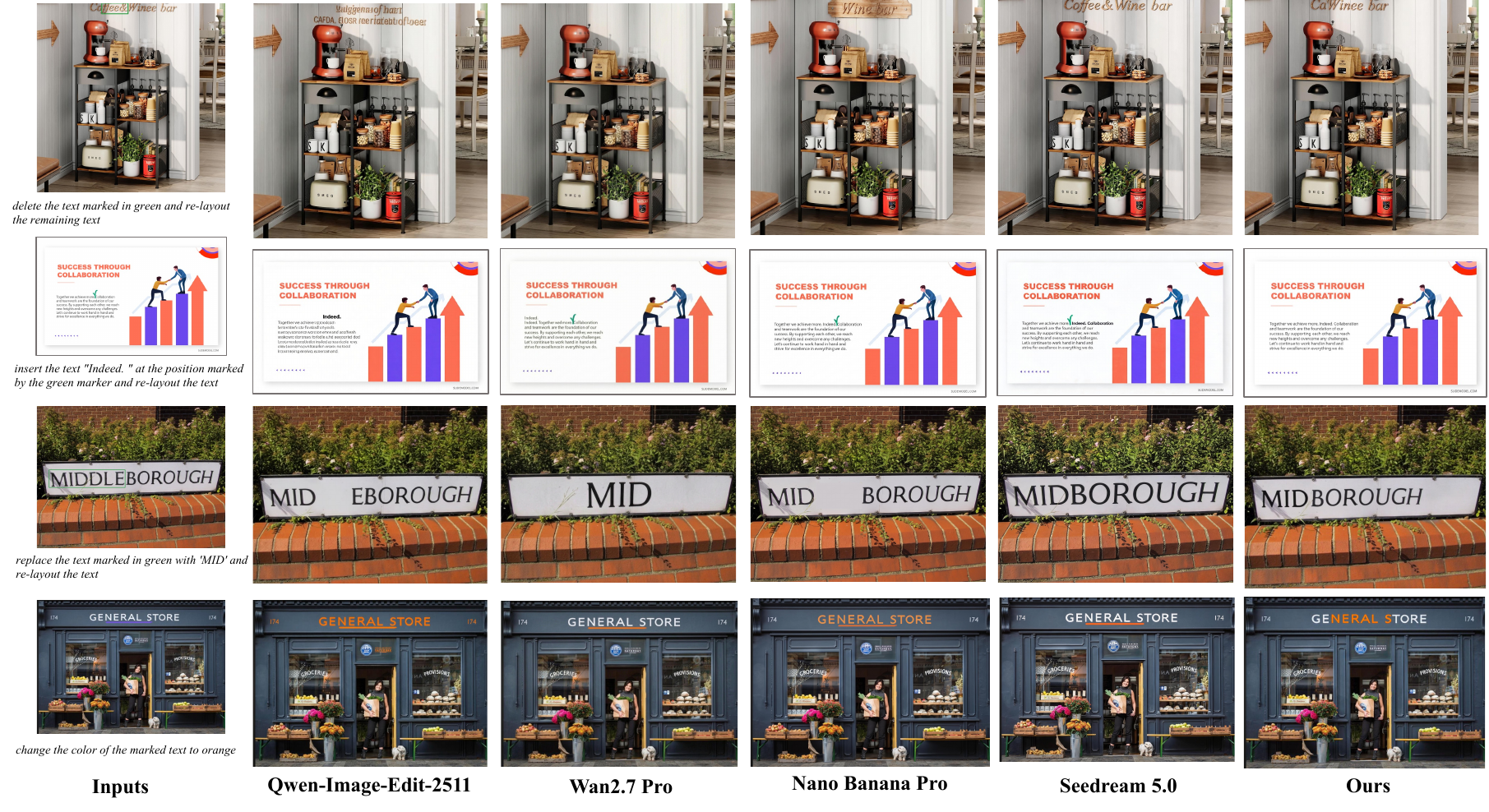}
  \caption{Qualitative examples for scribble-guided text editing.}
  \label{fig:supp-text-qualitative}
\end{figure}

\subsection{Qualitative Results of Text Editing}
\label{sec:supp:text:qualitative}

Figure~\ref{fig:supp-text-qualitative} shows representative VIBE-Text examples across text-editing tasks. The comparison is intended to highlight whether the model edits only the scribbled text span, preserves unrelated text, and performs local re-layout when the requested edit changes line length or typography.

\section{Limitations}
\label{sec:supp:limitation}

This work focuses on scribble-guided editing at the \emph{Deictic Level}, following the terminology introduced by VIBE. In this setting, the scribble gives an explicit spatial cue for the editing instruction, so the main challenge is to ground the marked region and perform the requested edit. This choice allows us to study spatial grounding and edit execution in a controlled way, but it also limits the scope of the paper. Higher-level instruction settings in VIBE require deeper visual understanding and reasoning: the model may need to infer object relations, recognize object states, resolve implicit references, or plan multi-step edits before it can decide where and how to edit. We leave these richer instruction forms to future work.

\section{Broader Impacts}
\label{sec:supp:broader-impacts}

Scribble-guided editing can make image editing more accessible by allowing users to specify spatial intent with a lightweight visual cue rather than detailed masks or expert tools. This may benefit creative workflows, rapid prototyping, educational content creation, and assistive editing interfaces where precise but easy-to-provide control is useful.

At the same time, our method inherits the broader risks of image editing and generative models. More controllable editing could be misused to alter images in misleading ways or to produce unwanted manipulations of visual content. These risks are not unique to our approach, but they motivate responsible release practices, such as documenting intended use cases, discouraging deceptive applications, and applying appropriate safeguards when deploying the model in user-facing systems.

%% file: sections/X_vibe_text_prompts.tex
\paragraph{Evaluation prompts.}
The following listings give the three VIBE-Text evaluation prompts used for $\mathcal{IA}$, $\mathcal{CP}$, and $\mathcal{VC}$.

\begin{lstlisting}[caption={VIBE-Text prompt for Instruction Adherence.},label={lst:vibe-text-ia}]
You are given THREE images and ONE text prompt.

The first image:
- This is the original image.

The second image:
- Visual instructions are drawn on this image based on the original image by the user.

The third image:
- This is the image generated by a model after editing the second image.

TEXT PROMPT:
- {prompt}

Your task is to evaluate whether the Output Image (The third image) correctly follows the instruction.
The instruction consists of BOTH:
(1) the visual instruction drawn on the second image, and
(2) the textual description in the Text Prompt.

You must independently evaluate the following FOUR metrics and assign a binary score (1 or 0) to each:

1. Visual Instruction Localization Correctness
Did the main edit occur on the text target explicitly indicated by the visual instruction on the Input Image (The second image)?

2. Visual Operator Type Compliance
Was the type of edit consistent with the operation implied by the visual instruction?

3. Textual Action Semantic Compliance
Did the model execute the core textual action specified in the Text Prompt?

4. Text Re-layout Compliance
When the requested text edit changes the amount, size, or placement of text, did the model adjust the affected line, paragraph, or text block layout appropriately?

- Judge whether the new layout looks reasonable and natural for the affected line, paragraph, or text block.
- The result should not leave obvious blank gaps, squeezed text, overlapping characters, or broken layout.

Scoring rules:
- Score 1 if the requirement is clearly satisfied.
- Score 0 if the requirement is not satisfied or is ambiguous.
- If unsure, assign 0.
- Partial compliance must be scored as 0.

You may reason freely to reach your decision.
Then, for EACH metric, provide:
- "score": an integer value of 0 or 1.
- "reason": ONE short factual sentence describing an observable outcome.
This summary must strictly follow the output format specified below:

{
  "Visual_Instruction_Localization_Correctness": {
    "reason": "brief factual summary",
    "score": 1/0
  },
  "Visual_Operator_Type_Compliance": {
    "reason": "brief factual summary",
    "score": 1/0
  },
  "Textual_Action_Semantic_Compliance": {
    "reason": "brief factual summary",
    "score": 1/0
  },
  "Text_Re-layout_Compliance": {
    "reason": "brief factual summary",
    "score": 1/0
  }
}
\end{lstlisting}

\begin{lstlisting}[caption={VIBE-Text prompt for Contextual Preservation.},label={lst:vibe-text-cp}]
You are given TWO images and ONE text prompt.

IMAGE 1 (Input Image):
- Original image with user-drawn visual instructions.

IMAGE 2 (Output Image):
- Image generated by the model after editing IMAGE 1.
- The output may be cropped or reframed.

TEXT PROMPT:
- {prompt}

Your task is to evaluate Contextual Preservation.

Definition:
Contextual Preservation checks whether the model changed anything it was NOT supposed to change.
It does NOT judge whether the edit was correct, precise, or well aligned.
Errors in edit location, extent, or alignment belong to Instruction Adherence, not Contextual Preservation.

Evaluation rules (follow strictly):

1) Cropping rule
- If the output is cropped, only compare the overlapping visible region.
- Ignore content missing only due to cropping.

2) Difference listing (what counts as a difference)
- List ONLY meaningful differences in:
  - text content (characters, words),
  - text order (line order, word order),
  - typography (font family, size, weight, color, emphasis, case),
  - non-text visual entities (objects, background).
- Do NOT list differences caused only by:
  - minor blur or softness,
  - small texture or color shifts,
  - pixel-level noise,
  - slight position or alignment offsets.

3) Target rule
- Identify the intended edit target based ONLY on:
  (a) the visual instruction marks, and
  (b) the text prompt.

4) Re-layout scope rule
- Reasonable line wrapping, spacing changes, and local repositioning INSIDE the affected text block are IN_TARGET when they are a consequence of the requested edit, even if nearby unedited words shift to new line positions inside that same affected block.
- Failed or awkward reflow inside the affected block is NOT judged here.
- This rule NEVER licenses changing unrelated text content, unrelated typography, or unrelated text blocks; those remain OUT_OF_TARGET.

5) Classification rule
- IN_TARGET:
  - any change within the intended target,
  - OR any imperfect attempt to edit the target (including misplacement, offset, scale error, or incomplete coverage).
- OUT_OF_TARGET:
  - any change to unrelated objects or regions,
  - any addition or removal of unrelated semantic entities,
  - any structural damage to non-target objects.

6) Scoring
- Score = 1 if NO OUT_OF_TARGET differences exist.
- Score = 0 if ANY OUT_OF_TARGET difference exists.
- If unsure, score = 0.

Output format:
First provide a brief analysis with these sections:
- ## Differences
- ## Target
- ## Classification
- ## Decision

Then output the final JSON as the last part of your response:

{
  "Text_Contextual_Preservation": {
    "reason": "string",
    "score": 0
  }
}
\end{lstlisting}

\begin{lstlisting}[caption={VIBE-Text prompt for Visual Coherence.},label={lst:vibe-text-vc}]
You are given THREE images and ONE text prompt.

The first image:
- This is the original image.

The second image
- Visual instructions are drawn on this image based on the original image by the user.

The third image:
- This is the image generated by a model after editing the second image.

TEXT PROMPT:
- {prompt}

Your task is to evaluate the Visual Coherence of the Output Image.

Visual Coherence evaluates whether the edited result is visually unified and generatively sound.
This metric is STYLE-AGNOSTIC: you must NOT judge realism, beauty, or artistic preference.
Instead, you must assess whether the edited image remains consistent with the source image and whether the output avoids clear generative artifacts.

You must independently evaluate the following THREE metrics and assign a binary score (1 or 0) to each:

1. Text Style Consistency
Did the edited text region adopt the same visual text style as the surrounding or corresponding source text (e.g., font family, font size, weight, color, perspective, rotation, curvature, and surface deformation)?

Scoring:
- Score 1 if the edited / added text clearly belongs to the same visual text domain and matches the surrounding source text style.
- Score 0 if it introduces a noticeably different from the surrounding source text.
- Do NOT judge whether the style looks good or realistic--only whether it matches the source domain.

2. Text Layout Seamlessness
Is the edited text visually integrated with the surrounding text line, paragraph, surface, or document layout, with no obvious local discontinuity?

Focus on clear local layout discontinuities such as:
- unnatural seams or hard boundaries around the edited text,
- overlapping characters or visibly broken character spacing within the edit,
- broken baseline, mismatched line height, or irregular word spacing in the immediate edited area,
- misalignment with the original text grid, sign surface, page margin, or perspective plane in the immediate edited area.

Scoring:
- Score 1 if the edited text integrates seamlessly with its immediate surrounding layout.
- Score 0 if there are clear and noticeable local layout discontinuities.

3. Artifact-Free Text Generation
Does the Output Image avoid obvious text-specific or general generative artifacts?

Consider artifacts such as:
- unreadable, garbled, melted, duplicated, or malformed glyphs in the edited text,
- broken or inconsistent strokes,
- ghost text or leftover erased text under the edit,
- unintended blur, pixelation, warping, or rendering collapse,
- residual visual instruction marks such as circles, underlines, or masks that should not appear in the final image.

Scoring:
- Score 1 if no clear generative artifacts are present in the edited result.
- Score 0 if obvious artifacts are visible.
- Style-specific noise or abstraction is allowed if consistent with the source domain.

General rules:
- Evaluate ONLY visual coherence, not instruction correctness or contextual preservation.
- Minor pixel-level differences are acceptable.
- If you are unsure about a case, assign score 0.
- Partial compliance must be scored as 0.

You may reason freely to reach your decision.
Then, for EACH metric, you provide a summary:
- "score": an integer value of 0 or 1.
- "reason": ONE short factual sentence describing an observable outcome.
This summary must strictly follow the output format specified below:

{
  "Text_Style_Consistency": {
    "reason": "brief factual summary",
    "score": 1/0
  },
  "Text_Layout_Seamlessness": {
    "reason": "brief factual summary",
    "score": 1/0
  },
  "Artifact-Free_Text_Generation": {
    "reason": "brief factual summary",
    "score": 1/0
  }
}
\end{lstlisting}